%% file: main.tex
\documentclass{article}

\usepackage[final, nonatbib]{nips_2017}
\usepackage[round]{natbib}

\usepackage[utf8]{inputenc} 
\usepackage[T1]{fontenc}    
\usepackage{hyperref}       
\usepackage{url}            
\usepackage{booktabs}       
\usepackage{amsfonts}       
\usepackage{nicefrac}       
\usepackage{microtype}      
\usepackage{color}
\usepackage{multirow}
\usepackage{tikz}
\usepackage[auth-lg]{authblk}
\usetikzlibrary{arrows.meta,calc,decorations.markings,math,arrows.meta}
\usepackage[]{subfigure}

\usepackage{amsmath}
\usepackage{bm}

\usepackage{floatrow}
\newfloatcommand{capbtabbox}{table}[][\FBwidth]
\newfloatcommand{capbfigbox}{figure}[][\FBwidth]

\renewcommand\cite{\citep} 
\newcommand\newcite{\citet}

\newcommand{\M}[2]{\bm{#1}_{#2}}
\newcommand{\V}[1]{\bm{#1}}

\newcommand{\argmax}{\operatornamewithlimits{argmax}}

\title{Recurrent Additive Networks}

\author{
Kenton Lee$^{\dagger}$\thanks{The first two authors contributed equally to this paper.}
 \qquad Omer Levy$^{\dagger*}$
 \qquad Luke Zettlemoyer$^{\dagger\ddagger}$\\

$^{\dagger}$Paul G. Allen School, University of Washington, Seattle, WA\\
$^{\ddagger}$Allen Institute for Artificial Intelligence, Seattle, WA\\
\texttt{\{kentonl,omerlevy,lsz\}@cs.washington.edu}

\qquad

}

\begin{document}

\maketitle

\begin{abstract}
We introduce \emph{recurrent additive networks} (RANs), a new gated RNN which is distinguished by the use of purely additive latent state updates. At every time step, the new state is computed as a gated component-wise sum of the input and the previous state, without any of the non-linearities commonly used in RNN transition dynamics. We formally show that RAN states are weighted sums of the input vectors, and that the gates only contribute to computing the weights of these sums. Despite this relatively simple functional form, experiments demonstrate that RANs perform on par with LSTMs on benchmark language modeling problems. This result shows that many of the non-linear computations in LSTMs and related networks are not essential, at least for the problems we consider, and suggests that the gates are doing more of the computational work than previously understood.
\end{abstract}

\section{Introduction}

Gated recurrent neural networks (GRNNs), such as long short-term memories (LSTMs)~\cite{lstm} and gated recurrent units (GRUs)~\cite{cho2014}, have become ubiquitous in natural language processing (NLP). GRNN's widespread popularity is at least in part due to their ability to model crucial language phenomena such as word order~\cite{adi2017}, syntactic structure~\cite{linzen2016}, and even long-range semantic dependencies~\cite{luhengsrl}. Like simple recurrent neural networks (S-RNNs) \cite{rnn}, they are able to learn non-linear functions of arbitrary-length input sequences, while at the same time alleviating the problem of vanishing gradients \cite{vangrad} by including gated additive state updates. While GRNNs work well in practice for a wide range of tasks, it is often difficult to interpret what function they have learned.





In this paper, we introduce a new GRNN architecture that is much simpler than existing approaches (e.g. fewer parameters and fewer non-linearities) and produces highly interpretable outputs, while matching the robust performance of LSTMs on benchmark language modeling tasks. More specifically, we propose \emph{recurrent additive networks} (RANs), which are distinguished by their use of purely additive latent state updates.
At every time step, the new state is computed as a gated component-wise sum of the input and the previous state. 
Unlike almost all existing RNNs, non-linearities affect the recurrent state only by controlling the gates at each timestep.


One advantage of simplifying the transition dynamics this way is that we can formally characterize the space of functions RANs compute. It is easy to show that the internal state of a RAN at each time step is simply a component-wise weighted sum of the input vectors up to that time.
Because all computations are component-wise, RANs can directly select which \emph{part} of each input element to retain at each time step, leading to a highly expressive yet interpretable model.

Despite their relative simplicity, RANs perform as well as LSTMs and related architectures on three language modeling benchmarks, but with far fewer parameters. To better understand this result, we 
derive the RAN updates from LSTM equations by (1) removing the recurrent non-linearity from $\widetilde{c_t}$ (LSTM's internal S-RNN) and (2) by removing the output gate.
Experiments show that we maintain the same level of performance after both simplifications, suggesting that additive connections, rather than the non-linear transition dynamics, are the driving force behind LSTM's success.

\section{Recurrent Additive Networks}
In this section, we first formally define the RAN model and then show that it represents a relatively simple class of additive functions over the input vectors.

\subsection{Model Definition}
\label{sec:model}
We assume a sequence of input vectors $\{\V{x}_1, \ldots, \V{x}_n\}$, and define a network that produces a sequence of output vectors $\{\V{h}_1, \ldots, \V{h}_n\}$. All recurrences over time are mediated by a sequence of state vectors $\{\V{c}_1, \ldots, \V{c}_n\}$, computed as follows for each time step $t$:
\begin{align}
\begin{split}
\widetilde{\V{c}}_{t} &= \M{W}{cx}\V{x}_t\\
\V{i}_{t}&=\sigma(\M{W}{ih} \V{h}_{t-1} + \M{W}{ix}\V{x}_t + \V{b}_{i})\\
\V{f}_{t}&=\sigma(\M{W}{fh}\V{h}_{t-1} + \M{W}{fx}\V{x}_t + \V{b}_{f})\\
\V{c}_{t}&=\V{i}_{t} \circ \widetilde{\V{c}}_{t}+\V{f}_{t} \circ \V{c}_{t-1}\\
\V{h}_{t}&=g(\V{c}_{t})
\label{eq:ran}
\end{split}
\end{align}
The new state $\V{c}_t$ is a weighted sum where two gates, $\V{i}_{t}$ (input) and $\V{f}_{t}$ (forget), control the mixing of the content layer $\widetilde{\V{c}}_{t}$ and the previous state $\V{c}_{t-1}$. The \emph{content layer} $\widetilde{\V{c}}_{t}$ is a linear transformation $\M{W}{x}$ over the input $\V{x}_{t}$, which is useful when the number of input dimensions $d_i$ is different from the number of hidden dimensions $d_h$ (e.g. embedding one-hot vectors). We also include an \emph{output layer} $\V{h}_{t}$ computed with a function $g$ of the internal state $\V{c}_{t}$. In our experiments, we use $g(x)=\tanh(x)$ and the identity function $g(x)=x$. The matrices $\M{W}{*}$ and biases $\V{b}_{*}$ are free trainable parameters.

The content layer $\widetilde{\V{c}}_{t}$ and output function $g$ are presented above to be consistent with other GRNN notation. However, the content layer $\widetilde{\V{c}}_{t}$ in RANs is very simple and only serves to allow different input vector and state vector dimensions. Similarly, the output function can be the identity, merging $\V{h}_{t}$ with $\V{c}_{t}$. When the content and output layers are trivial, the RAN can be reduced to an even simpler form:
\begin{align}
\begin{split}
\V{i}_{t}&=\sigma(\M{W}{ic} \V{c}_{t-1} + \M{W}{ix}\V{x}_t + \V{b}_{i})\\
\V{f}_{t}&=\sigma(\M{W}{fc}\V{c}_{t-1} + \M{W}{fx}\V{x}_t + \V{b}_{f})\\
\V{c}_{t}&=\V{i}_{t} \circ \V{x}_{t}+\V{f}_{t} \circ \V{c}_{t-1}
\label{eq:simple_ran}
\end{split}
\end{align}
Unlike LSTMs and GRUs, RANs only use additive connections to update the latent state $\V{c}_t$. RANs also use relatively fewer parameters. For example, the number of parameters (omitting biases) in an LSTM are $4d_h^2 + 4d_h d_i$, but only $2d_h^2 + 3d_h d_i$ in a RAN. In Section \ref{sec:deconstruction}, we explore their relationships more closely, showing that RANs are a significantly simplified variation of both LSTMs and GRUs. These simplifications, perhaps surprisingly, perform just as well as LSTMs on language modeling (Section~\ref{sec:results}). They also lead to a highly interpretable model that can be careful analyzed, as we present in more detail in the rest of this section and in Section~\ref{sec:visualization}.

\subsection{Analysis}

Another advantage of the relative simplicity of RANs is that we can formally characterize the space of functions that are used to compute the hidden states $\V{c}_t$. In particular, each state is a \emph{component-wise weighted sum of the input} with the form:
\begin{align}
\begin{split}
\V{c}_{t}&=\V{i}_{t} \circ \V{x}_{t}+\V{f}_{t} \circ \V{c}_{t-1}\\
&= \sum_{j=1}^{t} \Big(\V{i}_j \circ \prod_{k=j+1}^{t} \V{f}_k \Big) \circ \V{x}_{j}\\
&= \sum_{j=1}^{t} \V{w}_{j}^{t} \circ \V{x}_{j}
\label{eq:explicit_weights}
\end{split}
\end{align}
when considering the simpler RAN described in equation set (\ref{eq:simple_ran}).\footnote{The state is also a linear function of the inputs in the more general form (equation set (\ref{eq:ran})). However, it is a weighted sum of linearly transformed inputs, instead of a weighted sum of the input vectors themselves.}
Each weight $\V{w}_{j}^{t}$ is a product of the input gate $\V{i}_{j}$ (when its respective input $\V{x}_{j}$ was read) and every subsequent forget gate $\V{f}_{k}$. An interesting property of these weights is that, like the gates, they are also soft component-wise binary filters. This also produces a highly interpretable model, where each component of each state can be directly traced back to the inputs that contributed the most to its sum.

\section{Language Modeling Experiments}
\label{sec:results}

We compare the RANs' performance to LSTMs on three benchmark language modeling tasks. For each dataset, we use previously reported hyperparameter settings that were tuned for LSTMs in order to ensure a fair comparison (Section~\ref{subsec:setup}). Our experiments show that RANs perform on par with LSTMs on all three benchmarks (Section~\ref{subsec:results}). The code and settings to replicate these experiments is publicly available.\footnote{\url{http://www.github.com/kentonl/ran}}

\subsection{Experiment Setup}
\label{subsec:setup}

\paragraph{Penn TreeBank}
The Penn Treebank (PTB) \cite{ptb} is a popular language-modeling benchmark, containing approximately 1M tokens over a vocabulary of 10K words. We used the implementation of \newcite{zaremba2014} while replacing any invocation of LSTMs with RANs. We tested two configurations: \textit{medium}, which uses two layers of 650-dimension RANs, and \textit{large}, which uses two layers of 1500-dimension RANs. Word embedding size is set to match the recurrent layers' size, and dropout \cite{dropout} is used throughout the network. Both settings use stochastic gradient descent (SGD) to optimize the model, each with a unique hyperparameter setting to gradually decrease the learning rate.\footnote{The only modifications we made from the original setting was to use lower initial learning rates to improve stability. Exact values are provided in the code.}

\paragraph{Billion-Word Benchmark}
Google's billion-word benchmark (BWB) \cite{chelba2014} is about a thousand times larger than PTB, and uses a more diverse vocabulary of 800K words. Using the implementation of \newcite{lmlimits}, we tested the \textit{LSTM-2048-512} configuration, which uses a single-layered LSTM of 2048 hidden dimensions and a word embedding space of 512 dimensions. In our experiments, the LSTM was replaced with a RAN, while reusing exactly the same hyperparameters (dimensions, dropout, learning rates, etc) that were originally tuned for LSTMs by Jozefowicz et al. Following their implementation, we project the hidden state at each time step down to 512 dimensions. Due to the enormous size of this dataset, we stopped training after 5 epochs.

\paragraph{Text8}
We also evaluated on a character-based language modeling benchmark, Text8.\footnote{\url{http://mattmahoney.net/dc/textdata}} This dataset is made of the first 100M characters in English Wikipedia after some filtering process. The vocabulary contains only 27 characters (lowercase a--z and space). We adapted the implementation of \newcite{zilly2017} by taking hyperparameter settings from \newcite{chung2017}, which had an LSTM-like architecture. Their setting used 128-dimension character embeddings, followed by 3 LSTM layers of 1024, 1024, and 2048 dimensions respectively, which we adapted to RANs of the same dimensions. We present only tanh RANs for this benchmark because identity RANs were unstable with these hyperparameters.

\subsection{Results}
\label{subsec:results}

We first compare the performance of RANs to that of LSTMs on the word-based language modeling benchmarks, PTB (Table~\ref{tab:ptb_results}) and BWB (Table~\ref{tab:google_results}). In all configurations, the change in performance is below 1\% relative difference between LSTMs and tanh RANs. It appears that despite using less than two-thirds the parameters, RANs can perform on par with LSTMs. The BWB result is particularly interesting, because it demonstrates RANs' ability to to leverage big data like LSTMs.

Perhaps an even more remarkable result is the fact that identity RANs -- which, excluding the gates, compute a \textit{linear} function of the input -- are also performing comparably to LSTMs on the BWB. This observation begs the question: how is it possible that a simple weighted sum performs just as well as an LSTM? In Section \ref{sec:deconstruction}, we show that LSTMs (and similarly GRUs) are implicitly computing some form of component-wise weighted sum as well, and that this computation is key to their success.

Finally, we compare the performance of RANs to a variety of LSTM variants on the character-based language modeling task Text8 (Table~\ref{tab:text8_results}). While the different results vary both in model and in hyperparameters, the same trend in which RANs perform similarly to LSTMs emerges yet again.

\begin{table}[t]
    \centering
    \begin{tabular}{l l c c}
        \toprule
        \textbf{Configuration} & \textbf{Model} & \textbf{Perplexity} & \textbf{\# RNN Parameters} \\
        \midrule
        \multirow{3}{*}{Medium} & LSTM \cite{zaremba2014} & 82.7 & ~~6.77M \\
        & \textbf{tanh RAN} & 81.9 & ~~4.23M \\
        & \textbf{identity RAN} & 85.5 & ~~4.23M \\
        \midrule
        \multirow{3}{*}{Large} & LSTM \cite{zaremba2014} & 78.4 & 36.02M \\
        & \textbf{tanh RAN} & 78.5 & 22.52M \\
        & \textbf{identity RAN} & 83.2 & 22.52M \\
        \bottomrule
    \end{tabular}
    \caption{The performance of RAN and LSTM on the Penn TreeBank (PTB) benchmark, measured by perplexity. We also display the number of RNN parameters for comparison.}
    \label{tab:ptb_results}
\end{table}

\begin{table}[t]
    \centering
    \begin{tabular}{l c c}
        \toprule
        \textbf{Model} & \textbf{Perplexity} & \textbf{\# RNN Parameters} \\
        \midrule
        LSTM \cite{lmlimits} & 47.5 & 9.46M \\
        \textbf{tanh RAN} & 47.9 & 6.30M \\
        \textbf{identity RAN} & 47.2 & 6.30M \\
        \bottomrule
    \end{tabular}
    \caption{The performance of RAN and LSTM on Google's billion-word benchmark (BWB), measured by perplexity after 5 epochs. We also display the number of RNN parameters for comparison.}
    \label{tab:google_results}
\end{table}

\begin{table}[t]
    \centering
    \begin{tabular}{l c}
        \toprule
        \textbf{Model} & \textbf{BPC} \\
        \midrule
        td-LSTM \cite{zhang2016} & 1.49 \\
        MI-LSTM \cite{wu2016} & 1.44 \\
        mLSTM \cite{krause2017} & 1.40 \\
        \textbf{tanh RAN} & 1.38 \\
        BatchNorm LSTM \cite{cooijmans2017} & 1.36 \\
        LayerNorm HM-LSTM \cite{chung2017} & 1.29 \\
        \midrule
        RHN \cite{zilly2017} & 1.27 \\
        \bottomrule
    \end{tabular}
    \caption{The performance of RAN and recently-published LSTM variants on the Text8 character-based language modeling benchmark, measured by bits per character (BPC).}
    \label{tab:text8_results}
\end{table}

\section{The Role of Recurrent Additive Networks in Long Short-Term Memory}
\label{sec:deconstruction}

The need for LSTMs is typically motivated by the fact that they can ease the vanishing gradient problem found in simple RNNs (S-RNNs) \cite{vangrad,lstm}. By introducing cell states that are controlled by a set of gates, LSTMs enable shortcuts through which gradients can flow easily when learning with backpropagation. This mechanism enables learning of long-distance dependencies while preserving the expressive power of recurrent non-linearities provided by S-RNNs. 

Rather than viewing the gating mechanism as simply an auxiliary mechanism to address a \emph{learning} problem, we present an alternate of view of LSTMs that emphasizes the \emph{modeling} strengths of the gates. We argue that it is possible to reinterpret LSTMs as a hybrid of two other recurrent architectures: (1) S-RNNs and (2) RANs. More specifically, LSTMs can be seen as computing a content layer using an S-RNN, which is then aggregated into a weighted sum using a RAN (Section~\ref{subsec:hybrid}). To better understand this composition, we show how a RAN can be derived by simplifying an LSTM (Section~\ref{subsec:derivation}) or a GRU (Section~\ref{subsec:gru}).
Given our empirical observations in Section~\ref{sec:results}, which show that RANs perform comparably with LSTMs, it appears that the recurrent non-linearity provided by S-RNN is not required for language modeling, and that the RAN is in fact performing the heavy lifting. 


\subsection{LSTM as a Hybrid of S-RNN and RAN}
\label{subsec:hybrid}

We first demonstrate the two sub-components of LSTM by dissecting its definition:
\begin{align}
\begin{split}
    \widetilde{\V{c}}_{t}&=\tanh(\M{W}{ch}\V{h}_{t-1} + \M{W}{cx}\V{x}_t  + \V{b}_{c})\\
    \V{i}_{t}&=\sigma(\M{W}{ih}\V{h}_{t-1} + \M{W}{ix}\V{x}_t + \V{b}_{i})\\
    \V{f}_{t}&=\sigma(\M{W}{fh}\V{h}_{t-1} + \M{W}{fx}\V{x}_t + \V{b}_{f})\\
    \V{o}_{t}&=\sigma(\M{W}{oh} \V{h}_{t-1} + \M{W}{ox}\V{x}_t + \V{b}_{o})\\
    \V{c}_{t}&=\V{i}_{t}\circ \widetilde{\V{c}}_{t}+\V{f}_{t} \circ \V{c}_{t-1}\\
    \V{h}_{t}&=\V{o}_{t}\circ \tanh(\V{c}_{t})
\end{split}
\end{align}
We refer to $\widetilde{\V{c}}_{t}$ as the content layer, which like S-RNNs is a non-linear recurrent layer.
The cell state $\V{c}_t$ behaves like a RAN, using input and forget gates to compute a weighted sum of the current content layer $\widetilde{\V{c}}_{t}$ and the previous cell state $\V{c}_{t-1}$.
In fact, just as in RANs, we can express each cell state as a component-wise weighted sum of the all previous content layers:
\begin{align*}
\V{c}_{t}&=\V{i}_{t}\circ \widetilde{\V{c}}_{t}+\V{f}_{t} \circ \V{c}_{t-1}\\
&= \sum_{j=0}^{t}\Big(\V{i}_{j}\circ \prod_{k=j+1}^{t} \V{f}_k \Big) \circ \widetilde{\V{c}}_{j}\\
&= \sum_{j=0}^{t}\V{w}_{j}^{t} \circ \widetilde{\V{c}}_{j}
\end{align*}
However, unlike RANs, the content layer $\widetilde{c_j}$ depends on both the current input and the previous state cell state. Therefore, LSTMs cannot express the cell state as a weighted sum of the \emph{input} vectors.


%


\subsection{Deriving RAN from LSTM}
\label{subsec:derivation}

To derive an RAN from LSTM's equations, we perform two steps: (1) simplify the output layer by removing the output gate, and (2) simplify the content layer using only a linear projection of the input.

\paragraph{Simplifying the Output Layer}

In the first step, we simply remove the output gate, and abstract the output non-linearity $\tanh$ with $g$:
\begin{align*}
    \widetilde{\V{c}}_{t}&=\tanh(\M{W}{ch}\V{h}_{t-1} + \M{W}{cx}\V{x}_t  + \V{b}_{c})\\
    \V{i}_{t}&=\sigma(\M{W}{ih}\V{h}_{t-1} + \M{W}{ix}\V{x}_t + \V{b}_{i})\\
    \V{f}_{t}&=\sigma(\M{W}{fh}\V{h}_{t-1} + \M{W}{fx}\V{x}_t + \V{b}_{f})\\
    \V{c}_{t}&=\V{i}_{t}\circ \widetilde{\V{c}}_{t}+\V{f}_{t} \circ \V{c}_{t-1}\\
    \V{h}_{t}&=g(\V{c}_{t})
\end{align*}
This step follows previous architectural ablations of LSTM \cite{jozefowicz2015,greff2016}, which demonstrated that removing the output gate has limited effects. 

\paragraph{Simplifying the Content Layer}

We proceed to remove the embedded S-RNN from the LSTM by eliminating the non-linearity of the content layer $\widetilde{\V{c}}_{t}$ and its dependence on the previous output vector $\V{h}_{t-1}$. Specifically, we replace the original content layer with a simple linear projection of the input vector $\widetilde{\V{c}}_{t} = \M{W}{x}\V{x}_t$, resulting in the RAN architecture:
\begin{align*}
\widetilde{\V{c}}_{t}&=\M{W}{cx}\V{x}_t\\
\V{i}_{t}&=\sigma(\M{W}{ih}\V{h}_{t-1} + \M{W}{ix}\V{x}_t + \V{b}_{i})\\
\V{f}_{t}&=\sigma(\M{W}{fh}\V{h}_{t-1} + \M{W}{fx}\V{x}_t + \V{b}_{f})\\
\V{c}_{t}&=\V{i}_{t}\circ \widetilde{\V{c}}_{t}+\V{f}_{t} \circ \V{c}_{t-1}\\
\V{h}_{t}&=g(\V{c}_{t})
\end{align*}
This step emphasizes the most significant difference between RANs and other RNNs. The fact that removing the embedded S-RNN does not harm performance on our benchmarks in any significant way suggests that it is perhaps unnecessary, and that the gating mechanism has sufficient modeling capacity of its own for language modeling.

\subsection{Deriving RAN from GRU}
\label{subsec:gru}

A similar analysis can be applied to other gated RNNs such as GRUs, which are typically defined as follows:
\begin{align}
\begin{split}
\V{z}_{t}&=\sigma(\M{W}{zh}\V{h}_{t-1} + \M{W}{zx}\V{x}_t + \V{b}_{z})\\
\V{r}_{t} &=\sigma(\M{W}{rh} \V{h}_{t-1} + \M{W}{rx}\V{x}_t + \V{b}_{r})\\
\widetilde{\V{h}}_{t}&=\tanh(\M{W}{hh}(r \circ \V{h}_{t-1}) + \M{W}{hx}\V{x}_t + \V{b}_{h})\\
\V{h}_{t}&=\V{z}_{t} \circ \widetilde{\V{h}}_{t} + (\V{1} - \V{z}_{t})\circ \V{h}_{t-1}
\end{split}
\end{align}
For ease of discussion, we present the following non-standard but equivalent form of GRUs, which aligns better with the notation from LSTMs and RANs:
\begin{align}
\begin{split}
\widetilde{\V{c}}_{t}&=\tanh(\M{W}{ch}\V{h}_{t-1} + \M{W}{cx}\V{x}_t + \V{b}_{c})\\
\V{i}_{t}&=\sigma(\M{W}{ic}\V{c}_{t-1} + \M{W}{ix}\V{x}_t + \V{b}_{i})\\
\V{o}_{t}&=\sigma(\M{W}{oc} \V{c}_{t-1} + \M{W}{ox}\V{x}_t + \V{b}_{o})\\
\V{c}_{t}&=\V{i}_{t} \circ \widetilde{\V{c}}_{t}+(\V{1} - \V{i}_{t})\circ \V{c}_{t-1}\\
\V{h}_{t}&=\V{o}_{t}\circ \V{c}_{t}
\end{split}
\end{align}
In this form, the content layer $\widetilde{\V{c}}_{t}$ is equivalent to $\widetilde{\V{h}}_{t}$. The input gate $\V{i}_{t}$ is equivalent to $\V{z}_{t}$, and the output gate $\V{o}_t$ is equivalent to the reset gate $\V{r}_{t}$. In this form, the output vector at each time step is $c_t$. A subtle detail that enables this transformation is that GRUs can be seen as maintaining separate cell and output states, similar to LSTMs. However, GRUs compute the content layer with respect to the previous hidden state $\V{h}_{t-1}$ rather than the previous output $\V{c}_{t-1}$.

In our alternate form, it is clear that GRUs also accumulate weighted summations $\V{c}_{t}$ of previous content layers $\widetilde{\V{c}}_{t}$. The derivation of RAN from here involves two straightforward steps. First, the non-linear recurrence of the content layer $\widetilde{\V{c}}_{t}$ is replaced by a simple linear projection $\widetilde{\V{c}}_{t} = \M{W}{cx}\V{x}_{t}$, as we did for LSTMs. This step also makes $\V{h}_{t}$ redundant. Second, we repurpose the output gate $\V{o}_{t}$ as a forget gate by applying it to the previous cell state $\V{c}_{t-1}$ rather than the current cell state $\V{c}_{t}$:
\begin{equation*}
\V{c}_{t}=\V{i}_{t} \circ \widetilde{\V{c}}_{t} + \V{o}_{t}\circ \V{c}_{t-1}
\end{equation*}
This step results in a RAN with an identity output function $g$:
\begin{align*}
\widetilde{\V{c}}_{t}&=\M{W}{cx}\V{x}_t\\
\V{i}_{t}&=\sigma(\M{W}{ic}\V{c}_{t-1} + \M{W}{ix}\V{x}_t + \V{b}_{i})\\
\V{o}_{t}&=\sigma(\M{W}{oc}\V{c}_{t-1} + \M{W}{ox}\V{x}_t + \V{b}_{o})\\
\V{c}_{t}&=\V{i}_{t}\circ \widetilde{\V{c}}_{t}+\V{o}_{t} \circ \V{c}_{t-1}
\end{align*}
An interesting effect of the coupled gates in GRU is that these weights are normalized, i.e. the weights over all previous time steps sum to $\V{1}$. As a result, GRUs are computing weighted \emph{averages} rather than weighted \emph{sums} of their content layer. In development experiments, we found that this normalization hurt performance if added to the RAN architecture. 
However, it is closely related to another architecture that computes weighted averages: attention \cite{attention}. In fact, we can consider a GRU as a \emph{component-wise} attention mechanism over its previous content layers and a RAN to be an unnormalized component-wise generalization of attention, which can point to many different possible inputs independently.

\section{Weight Visualization}
\label{sec:visualization}
\input{visualization}

Given the relative interpretability of the RAN predictions, we can trace the importance of each component in each of the previous elements explicitly, as shown in equation set~(\ref{eq:explicit_weights}). In language modeling, we can also compute which previous word had the overall strongest influence in predicting the next word at time step $t$:
\begin{align}
v_t = \argmax_{j=1}^{t-1} \big(\max_{m=1}^{d_h}(\V{w}_j^t(m))\big)
\label{eq:viz}
\end{align}
In Figure~\ref{fig:weight-visualization}, we visualize this computation by drawing arrows from each word $t$ to its most influential predecessor (i.e. the previous word with the highest weighted component).
In these four example sentences, we see that RANs can recover long distance dependencies that make intuitive sense given the syntactic and semantic structure of the text. For example, verbs are related to their arguments, even when coordinated, and nouns in relative clauses depend on the noun they are modifying. This illustration provides only a glimpse into what the model is capturing, and perhaps future, more detailed visualizations that take the individual components into account can provide further insight into what RANs are learning in practice.


\section{Related Work}

Many variants of LSTMs~\cite{lstm} and alternative designs for more general GRNNs have been proposed. One such variant adds peephole connections between the cell state and the gates \cite{gers2000}. GRUs \cite{cho2014} decrease the number of parameters by coupling the input and forget gates and rewiring the gate computations (see Section~\ref{subsec:gru} for an alternative formulation). \newcite{greff2016} conducted an LSTM ablation study that probed the importance of each component independently, while others \cite{jozefowicz2015,zoph2017} took an automatic approach to the task of architecture design, finding additional variants of GRNNs. While we demonstrate that RANs can be seen as an ablation of LSTMs or GRUs, they are vastly simpler than the variants explored in the aforementioned studies. Specifically, this is the first study to remove the recurrent non-linearity (the embedded simple RNN) from such models.

Concurrently,~\newcite{lei2017} arrived at a similar space of models as RANs by deriving recurrent neural networks from string kernels, which they call kernel neural networks (KNNs). Our studies complement each other in two ways. First, Lei et al. show the connection between RANs/KNNs and kernels, while we show the connection to LSTMs and GRUs. Second, we conduct somewhat different experiments; we compare RANs to LSTMs in standard settings that were tuned for LSTMs, while Lei et al. show that RANs can achieve state-of-the-art performance when the network's design is more flexible and can be fitted with the latest advances in language modeling.

Several approaches represent sequences as weighted sums of their elements. Perhaps the most popular mechanism in NLP is attention \cite{attention}, which assigns a normalized scalar weight to each element (typically a word vector) as a function of its compatibility with an external element. The ability to inspect attention weights has driven the use of more interpretable neural models. Self-attention~\cite{cheng2016,parikh2016} extends this notion by computing intra-sequence attention.~\newcite{vaswani2017} further showed that state-of-the-art machine translation can be achieved using only self-attention, without the use of recurrent non-linearities found in LSTMs. Recently, \newcite{arora2017} proposed a theory-driven approach to assign scalar weights to elements in a bag of words. While these methods assign scalar weights to each input explicitly, RANs implicitly compute a component-wise weighted sum as a byproduct of a simple RNN state scheme.



\section{Conclusion}

We introduced recurrent additive networks (RANs), a type of gated RNNs that performs purely additive updates to its latent state. While RANs do not include the non-linear recurrent connections that are typically considered to be crucial for RNNs, they have remarkably similar performance to LSTMs on several language modeling benchmarks. RANs are also considerably more transparent than other RNNs; their limited use of non-linearities enables the state vector to be expressed as a component-wise weighted sum of previous input vectors. This also allows the individual impact of the sequence inputs on the state to be recovered explicitly, directly pointing to which factors are most influencing each part of the current state.

This work also sheds light on the inner workings of existing, more opaque models, primarily LSTMs and GRUs. We demonstrate that RAN-like components exist within LSTMs and GRUs. Furthermore, we provide empirical evidence that the use of recurrent non-linearities within those architectures is perhaps unnecessary for language modeling.

While we demonstrate the robustness of RANs on several language modeling benchmarks, it is an open question whether these findings will generalize across a wider variety of tasks. However, the results do suggest that it may be possible to develop related, relatively simple, additive gated RNNs that are better building blocks for a wide range of different neural architectures. We hope that our findings prove helpful in the design of future recurrent neural networks.




\subsection*{Acknowledgements}
The research was supported in part by DARPA under the DEFT program (FA8750-13-2-0019), the ARO (W911NF-16-1-0121), the NSF (IIS-1252835, IIS-1562364), gifts from Google, Tencent, and Nvidia, and an Allen Distinguished Investigator Award. We also thank Yoav Goldberg, Benjamin Heinzerling, Tao Lei, Luheng He, Aaron Jaech, Ariel Holtzman, and the UW NLP group for helpful conversations and comments on the work.

\bibliographystyle{plainnat}
\bibliography{references}

\end{document}

%% file: visualization.tex
\newcommand\word[3] {
\node[w] (#1) at (#2, 0) {#1};
\draw[-{Stealth[scale=1.0]}] ([yshift=8]#1.base) to [out=120,in=60] ([yshift=6]#3.base);
}

\newcommand\firstword[2] {
\node[w] (#1) at (#2, 0) {#1};
}

\begin{figure}
\centering
\scalebox{0.88}{
\begin{tikzpicture}[
  ->,
  thick, shorten >=1pt,
  >=latex,
  w/.style={%
    text height=1.5ex,
    text depth=.25ex,
    text width=11em,
    text centered,
    minimum height=4em,
    anchor=south
  }
]
\clip(-0.2,0) rectangle (12, 2.5);
\firstword{An}{0}
\word{earthquake}{1.2}{An}
\word{struck}{2.7}{earthquake}
\word{northern}{4}{earthquake}
\word{California}{5.6}{northern}
\word{killing}{7.2}{northern}
\word{more}{8.5}{killing}
\word{than}{9.5}{killing}
\word{50}{10.5}{killing}
\word{people}{11.5}{killing}
\end{tikzpicture}
}
\scalebox{0.88}{
\begin{tikzpicture}[
  ->,
  thick, shorten >=1pt,
  >=latex,
  w/.style={%
    text height=1.5ex,
    text depth=.25ex,
    text width=11em,
    text centered,
    minimum height=4em,
    anchor=south
  }
]
\clip(-0.2,0) rectangle (7.5, 2.8);
\firstword{He}{0}
\word{sits}{1}{He}
\word{down}{2}{sits}
\word{at}{3}{sits}
\word{the}{4}{sits}
\word{piano}{5}{sits}
\word{and}{6}{piano}
\word{plays}{7}{piano}
\end{tikzpicture}
}
\scalebox{0.88}{
\begin{tikzpicture}[
  ->,
  thick, shorten >=1pt,
  >=latex,
  w/.style={%
    text height=1.5ex,
    text depth=.25ex,
    text width=11em,
    text centered,
    minimum height=4em,
    anchor=south
  }
]
\clip(-1,0) rectangle (12, 2.5);
\firstword{Conservative}{0}
\word{party}{1.5}{Conservative}
\word{fails}{2.5}{Conservative}
\word{to}{3.2}{fails}
\word{secure}{4.0}{fails}
\word{a}{4.7}{secure}
\word{majority}{5.7}{a}
\word{resulting}{7.2}{a}
\word{in}{8.1}{resulting}
\word{a}{8.6}{resulting}
\word{hung}{9.3}{resulting}
\word{parliament}{10.7}{resulting}
\end{tikzpicture}
}
\caption{Partial visualizations of the weights in the weighted sum computed by an RAN ($\V{w}_j^t$ in equation set (\ref{eq:explicit_weights})). The arrows indicate for each word $t$ which previous word has the highest weighted component ($v_t$ in equation (\ref{eq:viz})).}
\label{fig:weight-visualization}
\end{figure}
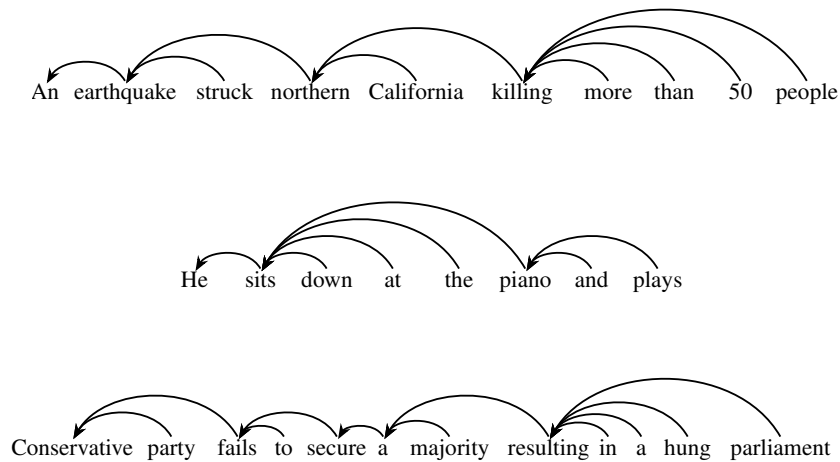

%% file: main.bbl
\begin{thebibliography}{28}
\providecommand{\natexlab}[1]{#1}
\providecommand{\url}[1]{\texttt{#1}}
\expandafter\ifx\csname urlstyle\endcsname\relax
  \providecommand{\doi}[1]{doi: #1}\else
  \providecommand{\doi}{doi: \begingroup \urlstyle{rm}\Url}\fi

\bibitem[Adi et~al.(2017)Adi, Kermany, Belinkov, Lavi, and Goldberg]{adi2017}
Yossi Adi, Einat Kermany, Yonatan Belinkov, Ofer Lavi, and Yoav Goldberg.
\newblock Fine-grained analysis of sentence embeddings using auxiliary
  prediction tasks.
\newblock In \emph{ICLR}, 2017.

\bibitem[Arora et~al.(2017)Arora, Liang, and Ma]{arora2017}
Sanjeev Arora, Yingyu Liang, and Tengyu Ma.
\newblock A simple but tough-to-beat baseline for sentence embeddings.
\newblock In \emph{ICLR}, 2017.

\bibitem[Bahdanau et~al.(2015)Bahdanau, Cho, and Bengio]{attention}
Dzmitry Bahdanau, Kyunghyun Cho, and Yoshua Bengio.
\newblock Neural machine translation by jointly learning to align and
  translate.
\newblock In \emph{ICLR}, 2015.

\bibitem[Bengio et~al.(1994)Bengio, Simard, and Frasconi]{vangrad}
Yoshua Bengio, Patrice~Y. Simard, and Paolo Frasconi.
\newblock Learning long-term dependencies with gradient descent is difficult.
\newblock \emph{IEEE Transactions on Neural Networks}, 5\penalty0 (2):\penalty0
  157--166, 1994.

\bibitem[Chelba et~al.(2014)Chelba, Mikolov, Schuster, Ge, Brants, and
  Koehn]{chelba2014}
Ciprian Chelba, Tomas Mikolov, Mike Schuster, Qi~Ge, Thorsten Brants, and
  Phillipp Koehn.
\newblock One billion word benchmark for measuring progress in statistical
  language modeling.
\newblock In \emph{INTERSPEECH}, 2014.

\bibitem[Cheng et~al.(2016)Cheng, Dong, and Lapata]{cheng2016}
Jianpeng Cheng, Li~Dong, and Mirella Lapata.
\newblock Long short-term memory-networks for machine reading.
\newblock In \emph{Proceedings of the 2016 Conference on Empirical Methods in
  Natural Language Processing}, pages 551--561, Austin, Texas, November 2016.
  Association for Computational Linguistics.
\newblock URL \url{https://aclweb.org/anthology/D16-1053}.

\bibitem[Cho et~al.(2014)Cho, van Merrienboer, Gulcehre, Bahdanau, Bougares,
  Schwenk, and Bengio]{cho2014}
Kyunghyun Cho, Bart van Merrienboer, Caglar Gulcehre, Dzmitry Bahdanau, Fethi
  Bougares, Holger Schwenk, and Yoshua Bengio.
\newblock Learning phrase representations using rnn encoder--decoder for
  statistical machine translation.
\newblock In \emph{Proceedings of the 2014 Conference on Empirical Methods in
  Natural Language Processing (EMNLP)}, pages 1724--1734, Doha, Qatar, October
  2014. Association for Computational Linguistics.
\newblock URL \url{http://www.aclweb.org/anthology/D14-1179}.

\bibitem[Chung et~al.(2017)Chung, Ahn, and Bengio]{chung2017}
Junyoung Chung, Sungjin Ahn, and Yoshua Bengio.
\newblock Hierarchical multiscale recurrent neural networks.
\newblock In \emph{ICLR}, 2017.

\bibitem[Cooijmans et~al.(2017)Cooijmans, Ballas, Laurent, and
  Courville]{cooijmans2017}
Tim Cooijmans, Nicolas Ballas, C{\'e}sar Laurent, and Aaron~C. Courville.
\newblock Recurrent batch normalization.
\newblock In \emph{ICLR}, 2017.

\bibitem[Elman(1990)]{rnn}
Jeffrey~L. Elman.
\newblock Finding structure in time.
\newblock \emph{Cognitive Science}, 14:\penalty0 179--211, 1990.

\bibitem[Gers and Schmidhuber(2000)]{gers2000}
Felix~A. Gers and J\"{u}rgen Schmidhuber.
\newblock Recurrent nets that time and count.
\newblock In \emph{IJCNN}, 2000.

\bibitem[Greff et~al.(2016)Greff, Srivastava, Koutn{\'\i}k, Steunebrink, and
  Schmidhuber]{greff2016}
Klaus Greff, Rupesh~K Srivastava, Jan Koutn{\'\i}k, Bas~R Steunebrink, and
  J{\"u}rgen Schmidhuber.
\newblock Lstm: A search space odyssey.
\newblock \emph{IEEE Transactions on Neural Networks and Learning Systems},
  2016.

\bibitem[He et~al.(2017)He, Lee, Lewis, and Zettlemoyer]{luhengsrl}
Luheng He, Kenton Lee, Mike Lewis, and Luke Zettlemoyer.
\newblock Deep semantic role labeling: What works and what’s next.
\newblock In \emph{Proceedings of the Annual Meeting of the Association for
  Computational Linguistics}, 2017.

\bibitem[Hochreiter and Schmidhuber(1997)]{lstm}
Sepp Hochreiter and J{\"u}rgen Schmidhuber.
\newblock {Long Short-term Memory}.
\newblock \emph{Neural computation}, 9\penalty0 (8):\penalty0 1735--1780, 1997.

\bibitem[J{\'o}zefowicz et~al.(2015)J{\'o}zefowicz, Zaremba, and
  Sutskever]{jozefowicz2015}
Rafal J{\'o}zefowicz, Wojciech Zaremba, and Ilya Sutskever.
\newblock An empirical exploration of recurrent network architectures.
\newblock In \emph{ICML}, 2015.

\bibitem[J{\'o}zefowicz et~al.(2016)J{\'o}zefowicz, Vinyals, Schuster, Shazeer,
  and Wu]{lmlimits}
Rafal J{\'o}zefowicz, Oriol Vinyals, Mike Schuster, Noam Shazeer, and Yonghui
  Wu.
\newblock Exploring the limits of language modeling.
\newblock \emph{arXiv preprint arXiv:1602.02410}, 2016.

\bibitem[Krause et~al.(2017)Krause, Murray, Renals, and Lu]{krause2017}
Ben Krause, Iain Murray, Steve Renals, and Liang Lu.
\newblock Multiplicative {LSTM} for sequence modelling.
\newblock In \emph{ICLR Workshop}, 2017.

\bibitem[Lei et~al.(2017)Lei, Jin, Barzilay, and Jaakkola]{lei2017}
Tao Lei, Wengong Jin, Regina Barzilay, and Tommi Jaakkola.
\newblock Deriving neural architectures from sequence and graph kernels.
\newblock In \emph{ICML}, 2017.

\bibitem[Linzen et~al.(2016)Linzen, Dupoux, and Goldberg]{linzen2016}
Tal Linzen, Emmanuel Dupoux, and Yoav Goldberg.
\newblock Assessing the ability of lstms to learn syntax-sensitive
  dependencies.
\newblock \emph{TACL}, 4:\penalty0 521--535, 2016.

\bibitem[Marcus et~al.(1993)Marcus, Santorini, and Marcinkiewicz]{ptb}
Mitchell~P. Marcus, Beatrice Santorini, and Mary~Ann Marcinkiewicz.
\newblock Building a large annotated corpus of english: The penn treebank.
\newblock \emph{Computational Linguistics}, 19:\penalty0 313--330, 1993.

\bibitem[Parikh et~al.(2016)Parikh, T\"{a}ckstr\"{o}m, Das, and
  Uszkoreit]{parikh2016}
Ankur Parikh, Oscar T\"{a}ckstr\"{o}m, Dipanjan Das, and Jakob Uszkoreit.
\newblock A decomposable attention model for natural language inference.
\newblock In \emph{Proceedings of the 2016 Conference on Empirical Methods in
  Natural Language Processing}, pages 2249--2255, Austin, Texas, November 2016.
  Association for Computational Linguistics.
\newblock URL \url{https://aclweb.org/anthology/D16-1244}.

\bibitem[Srivastava et~al.(2014)Srivastava, Hinton, Krizhevsky, Sutskever, and
  Salakhutdinov]{dropout}
Nitish Srivastava, Geoffrey~E. Hinton, Alex Krizhevsky, Ilya Sutskever, and
  Ruslan Salakhutdinov.
\newblock Dropout: A simple way to prevent neural networks from overfitting.
\newblock \emph{Journal of Machine Learning Research}, 15:\penalty0 1929--1958,
  2014.

\bibitem[Vaswani et~al.(2017)Vaswani, Shazeer, Parmar, Uszkoreit, Jones, Gomez,
  Kaiser, and Polosukhin]{vaswani2017}
Ashish Vaswani, Noam Shazeer, Niki Parmar, Jakob Uszkoreit, Llion Jones,
  Aidan~N. Gomez, Lukasz Kaiser, and Illia Polosukhin.
\newblock Attention is all you need.
\newblock \emph{arXiv preprint arXiv:1706.03762}, 2017.

\bibitem[Wu et~al.(2016)Wu, Zhang, Zhang, Bengio, and Salakhutdinov]{wu2016}
Yuhuai Wu, Saizheng Zhang, Ying Zhang, Yoshua Bengio, and Ruslan Salakhutdinov.
\newblock On multiplicative integration with recurrent neural networks.
\newblock In \emph{NIPS}, 2016.

\bibitem[Zaremba et~al.(2014)Zaremba, Sutskever, and Vinyals]{zaremba2014}
Wojciech Zaremba, Ilya Sutskever, and Oriol Vinyals.
\newblock Recurrent neural network regularization.
\newblock \emph{arXiv preprint arXiv:1409.2329}, 2014.

\bibitem[Zhang et~al.(2016)Zhang, Wu, Che, Lin, Memisevic, Salakhutdinov, and
  Bengio]{zhang2016}
Saizheng Zhang, Yuhuai Wu, Tong Che, Zhouhan Lin, Roland Memisevic, Ruslan
  Salakhutdinov, and Yoshua Bengio.
\newblock Architectural complexity measures of recurrent neural networks.
\newblock In \emph{NIPS}, 2016.

\bibitem[Zilly et~al.(2017)Zilly, Srivastava, Koutn{\'i}k, and
  Schmidhuber]{zilly2017}
Julian~G. Zilly, Rupesh~Kumar Srivastava, Jan Koutn{\'i}k, and J\"{u}rgen
  Schmidhuber.
\newblock Recurrent highway networks.
\newblock In \emph{ICLR}, 2017.

\bibitem[Zoph and Le(2017)]{zoph2017}
Barret Zoph and Quoc~V Le.
\newblock Neural architecture search with reinforcement learning.
\newblock In \emph{ICLR}, 2017.

\end{thebibliography}
